\documentclass[journal,twoside,web]{ieeecolor}
\usepackage{tmi}
\usepackage{cite}
\usepackage{amsmath,amssymb,amsfonts}
\usepackage{algorithmic}
\usepackage[ruled,vlined]{algorithm2e}
\usepackage{graphicx}
\usepackage{textcomp}
\usepackage[caption=false,font=footnotesize]{subfig}
\usepackage{multirow}
\usepackage{booktabs}

\def\BibTeX{{\rm B\kern-.05em{\sc i\kern-.025em b}\kern-.08em
    T\kern-.1667em\lower.7ex\hbox{E}\kern-.125emX}}
\begin{document}
\title{Adaptive Label Error Detection: A Bayesian Approach to Mislabeled Data Detection}
\author{Zan Chaudhry, Noam H. Rotenberg, Brian Caffo, Craig K. Jones, and Haris I. Sair
\thanks{Received 14th January 2026. \textit{(Corresponding authors: Craig K. Jones and Haris I. Sair.)}}
\thanks{Z. Chaudhry and N. H. Rotenberg were with the Department of Biomedical Engineering, Johns Hopkins University, Baltimore, MD 21218 USA. Z. Chaudhry is now with the Systems, Synthetic, and Quantitative Biology Program, Harvard University, Cambridge, MA 02138 USA (e-mail: zanchaudhry@fas.harvard.edu). N. H. Rotenberg is now with the Yale School of Medicine, New Haven, CT 06510 USA (e-mail: noam.rotenberg@yale.edu).}
\thanks{B. Caffo is with the Department of Biostatistics, Johns Hopkins Bloomberg School of Public Health, Baltimore, MD 21205 USA (e-mail: bcaffo1@jhu.edu)}
\thanks{C. K. Jones is with the Department of Computer Science, Johns Hopkins University, Baltimore, MD 21218 USA, and the Department of Radiology and Radiological Science, Johns Hopkins University School of Medicine, Baltimore MD 21287 USA (e-mail: craigj@jhu.edu).}
\thanks{H. I. Sair is with the Department of Radiology and Radiological Science, Johns Hopkins University School of Medicine, Baltimore MD 21287 USA, and the Malone Center for Engineering in Healthcare, The Whiting School of Engineering, Johns Hopkins University, Baltimore, MD 21218 USA (email: hsair1@jhmi.edu).}}

\maketitle

\begin{abstract}
Machine learning classification systems are susceptible to poor performance when trained with incorrect ground truth labels, even when data is well-curated by expert annotators. As machine learning becomes more widespread, it is increasingly imperative to identify and correct mislabeling to develop more powerful models. In this work, we motivate and describe Adaptive Label Error Detection (ALED), a novel method of detecting mislabeling. ALED extracts an intermediate feature space from a deep convolutional neural network, denoises the features, models the reduced manifold of each class with a multidimensional Gaussian distribution, and performs a simple likelihood ratio test to identify mislabeled samples. We show that ALED has markedly increased sensitivity---without compromising precision---compared to established label error detection methods, on multiple medical imaging datasets. We demonstrate an example where fine-tuning a neural network on corrected data results in a 33.8\% decrease in test set errors, providing strong benefits to end users. The ALED detector is deployed in the Python package \texttt{statlab}.
\end{abstract}

\begin{IEEEkeywords}
Deep Convolutional Neural Networks, Feature Space Geometry, Label Noise, Medical Image Classification, Mislabeled Data Detection
\end{IEEEkeywords}

\section{Introduction}
\label{sec:introduction}
\IEEEPARstart{M}{achine} learning (ML) and artificial intelligence (AI) tools have revolutionized every aspect of scientific research, and the field of medical imaging has both produced and benefited from many of these constant advances. In particular, deep convolutional networks (DCNNs) demonstrate tremendous potential in image segmentation and classification tasks for virtually all imaging modalities, in some cases exceeding the performance of trained physicians \cite{Anwar2018, Sarvamangala2021, esteva2017dermatologist, gulshan2016development, mckinney2020international}.

In supervised learning schemes, DCNNs are trained numerically using variations of the gradient-descent algorithm to minimize a loss criterion, which expresses a distance between model prediction and ground-truth labels for a particular set of inputs. The ground-truth labels determine the direction of model parameter tuning and thus the final model performance. However, human annotators are imperfect, and mislabeled data---where the recorded “ground truth” label of a sample is incorrect---is a serious concern in AI/ML research. Significant mislabeling has even been identified in highly-curated benchmark datasets, including MNIST, CIFAR, and ImageNet \cite{northcutt2021pervasivelabelerrorstest}. In the medical imaging domain, radiologist error is not trivial, with estimated day-to-day error rates of 3-5$\%$, and inter-observer discrepancy rates near 25$\%$ in certain radiological tasks \cite{Brady2016}. Based on the pervasiveness of labeling error, it is likely that most, if not all large datasets will have mislabeled samples. The state of AI/ML research would benefit from improved models trained on more accurately labeled data; however, expert annotation is costly, particularly in the medical imaging space. Thus, a number of algorithmic methods have been devised to train models in the presence of mislabeled samples. These methods generally fall into two classes: model-centric and data-centric approaches.

Model-centric approaches aim to improve the model’s robustness to label noise by training the model on noisy data and adjusting the model or training scheme accordingly (e.g., adapting the loss function). Some successful examples include: the method of Reed et al., which reweights the loss function according to the concurrence between the given noisy label and predicted true label, based on the consistency between likewise labeled samples in the feature space \cite{reed2014training}; the method of Goldberger et al., which adds an additional, learned layer to a neural network to model a confusion matrix between soft estimates of the true label and the given noisy label \cite{goldberger2017training}; Symmetric Cross Entropy Loss (SCELoss), which is an alternative loss function based on KL-divergence that aims to balance class-biased training, a condition in which certain classes may be easier to learn than others \cite{wang2019symmetric}; and Mentor-Net, which trains an additional “mentor” model to define a curriculum used to train the predictive model that prioritizes presenting samples with high label certainty \cite{jiang2018mentornet}.

Data-centric approaches, in contrast, aim to improve the dataset quality by measuring label noise and adjusting the dataset accordingly. Some successful examples include: Mixup, in which the model is trained on convex combinations of samples as a form of data augmentation, thus improving generalization behavior and preventing memorization of noisy labels; Iterative Noisy Cross Validation (INCV), in which models trained on cross-validation folds of the noisy dataset are applied to the other folds to identify mislabeled examples and iteratively remove them; and confident learning methods, which estimate the joint distribution between noisy and true labels using probabilistic thresholds on the model predictions, to identify mislabeled examples and then remove them \cite{zhang2017mixup, chen2019understanding, northcutt2021confident}. 

Confident learning, especially in its generalized form implemented by Cleanlab (CL), has become a popular tool due to its strong average-case performance. CL provides two confident learning approaches, the model predictions approach previously described, and a related method (“CL with features”) that operates in the model's intermediate feature space, using KNN clustering followed by the same confident learning steps. However, in practice, we observe that CL and CL with features prioritize precision at significant expense to recall in label error detection (see Section III). This results in low sensitivity, meaning many mislabeled samples remain undetected, particularly when classifier performance is degraded by label noise.

In this paper, we present Adaptive Label Error Detection (ALED), a data-centric approach that improves upon CL's recall in practical settings. ALED leverages the feature embeddings of a network trained on noisy data, then denoises the feature space and fits a Gaussian model to identify outlier embeddings likely corresponding to mislabeled samples.

 ALED has improved sensitivity and F1 score on mislabeled sample detection, which is valuable in realistic use cases where thresholding decisions are made without access to ground truth labels. We evaluate ALED across diverse datasets, model architectures, and noise conditions, showing that it robustly identifies mislabeled samples even when classifier predictions are unreliable. We also demonstrate the utility of ALED by training models on data cleaned with ALED, achieving significant performance gains over existing methods. Furthermore, we provide theoretical justification to support ALED’s design choices.

\section{Mathematical Motivation}

We begin by introducing the general steps that motivate the ALED algorithm design. First, given a set of training data and a classification task, we naturally desire a method to process the input data into a format that contains representative features that can be used to separate classes. In the case of complex, high-dimensional feature representations, it will be necessary to denoise and reduce the dimensionality of these feature data to a manageable level. Finally, we must find patterns in this low-dimensional space that allow us to identify mislabeled data points. 

We implement this workflow with DCNNs. DCNNs possess a powerful ability to uncover patterns in samples and find differences between classes, even in the presence of mislabeling \cite{rolnick2017deep}. However, outlier detection applied to DCNN predictions alone is generally insufficient to detect mislabeling for multiple reasons. In particular, DCNNs are suscpetible to overfitting to mislabeling, and there is significant information loss from intermediate feature representations to output predictions \cite{arpit2017closer}. We posit that for models trained with mislabeling, this intermediate feature space still generally contains sufficient information to separate the manifolds of the true underlying classes; the performance of our method and previous work supports this hypothesis \cite{lee2018simple}. Therefore, we extract this intermediate representation, denoise and reduce dimensionality, and utilize robust, ``low-power” methods to identify mislabeled samples while preventing overfitting. We explore the specifics and further justify this broad road map in the following sections.

\subsection{Definitions}
First, we define a neural network as a composition of weight matrices, $\mathbf{W}$, with bias terms $\mathbf{b}$, and element-wise nonlinear activation functions, $\sigma$. If a layer at depth $l$ receives an input, $\mathbf{x}^l$, the output is given by

\begin{equation}
    \mathbf{x}^{l+1} = \sigma \left( \mathbf{W}^l \mathbf{x}^l + \mathbf{b}^l \right)
\end{equation}

Then the neural network of depth $L$ with input data $\mathbf{x}$ is given by:

\begin{equation}
    \mathbf{f}(\mathbf{x}) = \sigma \left( \mathbf{W} ^{L}\left(\dots \sigma \left( \mathbf{W}^1 \mathbf{x} + \mathbf{b}^1 \right) + \dots \right) + \mathbf{b}^{L} \right)
\end{equation}

We also define the pre-activation of a given layer, $\mathbf{z}^l$ as:

\begin{equation}
    \mathbf{z}^{l} =  \mathbf{W}^{l-1} \mathbf{x}^{l-1} + \mathbf{b}^{l-1} 
\end{equation}

Although the defined network is fully-connected (FCNN), DCNNs can be represented in this form by reparameterizing convolutions as Toeplitz matrices. Additionally, we refer to the final layer that produces the model output as the classifier, and we refer to the intermediate layers as hidden layers. We define the term feature space to represent the outputs and pre-activations of the hidden layers.

The notion of the feature space spans many levels of detail. Similar to human visual processing, at different depths, DCNNs tend to learn different representations of the input data that become progressively more abstract and high-dimensional \cite{zeiler2014visualizing}. In this paper, we simply take the final hidden layer pre-activations ($\mathbf{z}^{L-1}$) to constitute the feature space (i.e. the pre-activations just before the classifier) \cite{donahue2014decaf}.

\subsection{Training with Label Noise}
Previous work has demonstrated that DCNNs can learn representative features of input data, even in the presence of mislabeling \cite{rolnick2017deep}. However, DCNNs are also often over-parameterized and can easily overfit/memorize training examples. This introduces the challenge with using model outputs for mislabeled data detection.

As a first consideration, there is significant information loss from the feature space to the output prediction space, since for wide networks, a very high-dimensional vector (the features) is transformed into a scalar or low-dimensional vector (the output). This alone is not a serious concern as the goal naturally of any classifier is to perform a dimensionality reduction: extracting a single prediction from this feature space. The challenge comes in the presence of mislabeling, when the weight vector (or matrix) may be overfit to label noise even while the feature space contains representative information. Significant evidence supports the notion that later layers of a DCNN are more likely to overfit \cite{stephenson2021geometrygeneralizationmemorizationdeep, zhang2021understanding}. 

The simplest explanation comes from gradient magnitude arguments during backpropagation (the same arguments for the vanishing gradient problem). Activation function derivatives generally have a magnitude less than 1, and so products of these derivatives during backpropagation naturally lead to a gradient magnitude reduction for early layers. This acts as an implicit stabilizer that allows early layers to be robust to noisy gradient updates, but the result is that later layers are more responsive to these gradient updates, increasing overfitting likelihood. Such gradient magnitude decreases through layers are observed consistently in experimental work \cite{glorot2010understanding}. Thus, we return to the intermediate features at some depth of the neural network for mislabeled data detection. The choice of feature space depth is a tradeoff between robustness to overfitting (using the earlier layers) and capturing rich feature information (using the deeper layers).

\subsection{Describing the Feature Space}
As discussed in section II.A, we define the feature space for DCNNs as the pre-activation of the final convolutional layer. We choose to operate in the feature space, as it (1) provides significantly more information and (2) is less prone to overfitting than the model output. To use this data, we first turn to an important result in the limit of infinite-width nerual networks. Previous work has demonstrated that infinite width neural networks are Gaussian processes, and that the pre-activations of wide (though finite-width) DCNNs are approximately Gaussian distributed \cite{lee2018deep}. We apply this approximation to the feature space of a binary classification problem, where we have classes $C=0$ and $C=1$. Thus, we assume the following distributions: $\mathbf{z}_{C=0}^{L-1} \sim \mathcal{N}\left( \boldsymbol{\mu}_{0}, \boldsymbol{\Sigma}_0 \right)$ and $\mathbf{z}_{C=1}^{L-1} \sim \mathcal{N}\left( \boldsymbol{\mu}_{1}, \boldsymbol{\Sigma}_1 \right)$.

Now we have a description of the feature space, and a notion of the class structure in this space. In practice (in the case of a two-dimensional input image), $\mathbf{z}^{L-1}$ is represented as $\mathbf{Z} \in \mathbb{R}^{N \times W \times W}$, a multidimensional array of feature maps. This is equivalent to a vector of dimension $d=N W^2$. For high $d$, traditional clustering methods will suffer from ``the curse of dimensionality." This issue cannot be completely avoided, but we mitigate this by first performing a dimensionality reduction: average pooling of the feature maps, reducing dimensionality to $d=N$. Average pooling preserves the normality assumption, as a linear combination of Gaussians will also be Gaussian. Furthermore, average pooling is implemented as a final dimensionality reduction in several DCNN architectures, prior to the classifier \cite{he2016deep, huang2017densely}.

Now we wish to further reduce dimensionality, as for several common architectures, $N$ is on the order of $10^3$ \cite{he2016deep, huang2017densely}. Based on empirical data, we found that a robust projection in the low dimensional space includes (i) a projection to maximize separability of the distributions $\mathbf{z}_0^{L-1}$ and $\mathbf{z}_1^{L-1}$ along the vector between the means and (ii) random projections from the high dimensional space. We motivate the first projection in Appendix I, showing that projecting the feature data along $\mathbf{v} = \boldsymbol{\mu}_{1} - \boldsymbol{\mu}_{0}$ approximately maximizes separability in high-dimensional settings. Additional information is captured by random projections, since by the Johnson-Lindenstrauss lemma, a set of orthogonal random projections will approximately conserve the structure of the data, without bias to potentially noise-corrupted directions, as would be the case with for example PCA \cite{johnson1984extensions}. In a high-dimensional space, we need not even enforce the orthogonality constraint, as any small set of random vectors is very likely to be approximately orthogonal \cite{blum2019foundations}.

To construct Gaussian clusters, we need an estimator of the covariance matrix. A natural choice is to use the given class labels (some of which may be incorrect) with a robust estimator, particularly the Minimum Covariance Determinant (MCD) method, which is robust to outliers. In the limit of infinite data points, the MCD method possesses a maximal breakdown value of 50$\%$ \cite{Hubert2017}. So finally, we have our extracted denoised/dimensionality-reduced feature vectors (we refer to these as $\mathbf{z}_{\text{reduced}}$) to which we have fit Gaussians by the MCD method. Given the inherent randomness introduced from projecting, we can further improve robustness by ensembling over multiple MCD fits on different random projections.

\subsection{Mislabeled Data Detection}
 If we view the problem as a simple binary classification task between our two robustly estimated Gaussian distributions, then the Bayes optimal test is the likelihood ratio test \cite{neyman1933ix}. Under this test, we use the probability density function of each class distribution ($f_{0}$ and $f_{1}$) and compare the likelihood of each sample's feature vector, $\mathbf{z_{\text{reduced}}}$ under each distribution to decide which distribution the vector belongs to. In our mislabeling context, we simply take the ratio of the likelihood under the alternative label and the likelihood under the reference label. Or, more explicitly:
\begin{equation}
    F(\mathbf{z}_{\text{reduced}})=\begin{cases} \text{mislabel} \text{  if } \Lambda(\mathbf{z}_{\text{reduced}})>\tau\\ \text{correct} \text{  if } \Lambda(\mathbf{z}_{\text{reduced}})\leq \tau\end{cases}
\end{equation}
where
\begin{equation}
    \Lambda(\mathbf{z_{\text{reduced}}}) = \frac{f_{\text{alt}}(\mathbf{z_{\text{reduced}}})}{f_{\text{given}}(\mathbf{z_{\text{reduced}}})}
\end{equation}
and $\tau$ is some threshold. This provides a statistically grounded thresholding method with strong theoretical guarantees on performance. This thesholding scheme is also the main gap between our method and CL/CL with features. 

\subsection{Algorithm Design}

\begin{algorithm}[t!]
\caption{\textbf{Adaptive Label Error Detection (ALED).} An ensemble-based approach using Minimum Covariance Determinant (MCD) on projected features to identify label noise.}
\label{ALED_alg}

\KwIn{\\
    \quad $X_{\textit{tr}}, y_{\textit{tr}}$: training features and potentially noisy binary labels\\
    \quad $\mathcal{M}$: deep convolutional neural network model\\
    \quad $\tau$: threshold for the likelihood ratio test\\
    \quad $r$: target dimensionality for random projections\\
    \quad $k$: ensemble size for likelihood estimation\\
\textbf{Requirement:}\\
    \quad $r < \text{len}(y_\text{tr})$}
\KwOut{\\
    \quad $\mathcal{M}_{\text{adj}}$: neural network $\mathcal{M}$ refined on cleaned labels}
\BlankLine
\textbf{Procedure:}\\
\tcc{Extract penultimate features}
$m \gets \text{len}(y_\text{tr})$\;
$\mathbf{Z}_{\text{pool}} \gets \big[ \mathbf{z}_1 | \mathbf{z}_2 | \dotsi | \mathbf{z}_m \big]$, where $\mathbf{z}_i$ is the average-pooled penultimate feature map of $\mathcal{M}(X_{\text{tr}}[i])$\;

Compute class centroids $\boldsymbol{\mu}_0, \boldsymbol{\mu}_1$ from $\mathbf{Z}_{\text{pool}}$ based on current $y_{\text{tr}}$\;

Initialize accumulated likelihoods $\mathbf{L}_{\text{given}} \gets \vec{0}, \mathbf{L}_{\text{alt}} \gets \vec{0}$\;

\BlankLine
\tcc{Ensemble Loop}
\For{$j \gets 1$ \KwTo $k$}{
    Generate $\mathbf{R} \in \mathbb{R}^{d \times r}$, a set of random projections\;
    Define projection matrix $\mathbf{W} \gets \begin{bmatrix} (\boldsymbol{\mu}_1 - \boldsymbol{\mu}_0)^\text{T } \\ \mathbf{R}^{\text{T }} \end{bmatrix}$\;
    $\mathbf{Z}_{\text{reduced}} \gets \mathbf{W} \mathbf{Z}_{\text{pool}}$ \tcp*{Project data}
    
    $(\hat{\boldsymbol{\mu}}_{0}, \hat{\boldsymbol{\mu}}_{1}, \hat{\boldsymbol{\Sigma}}_{0}, \hat{\boldsymbol{\Sigma}}_{1}) \gets$ MCD estimation\;
    
    \ForEach{sample $i \in [1, m]$}{
        $c_{\text{given}} \gets y_{\text{tr}}[i]$ \tcp*{Given class label}
        $c_{\text{alt}} \gets \neg y_{\text{tr}}[i]$ \tcp*{Alternate label}
        
        $f_{\text{given}} \gets f_\mathcal{N}\left(\mathbf{Z}_{\text{reduced}}[:,i] \mid \hat{\boldsymbol{\mu}}_{c_{\text{given}}}, \hat{\boldsymbol{\Sigma}}_{c_{\text{given}}}\right)$\;
        $f_{\text{alt}} \gets f_\mathcal{N}\left(\mathbf{Z}_{\text{reduced}}[:,i] \mid \hat{\boldsymbol{\mu}}_{c_{\text{alt}}}, \hat{\boldsymbol{\Sigma}}_{c_{\text{alt}}}\right)$\;
        
        $\mathbf{L}_{\text{given}}[i] \gets \mathbf{L}_{\text{given}}[i] + f_{\text{given}}$\;
        $\mathbf{L}_{\text{alt}}[i] \gets \mathbf{L}_{\text{alt}}[i] + f_{\text{alt}}$\;
    }
}

\BlankLine
\tcc{Label Correction}
\ForEach{sample $i \in [1, m]$}{
    $\bar{f}_{\text{given}} \gets \mathbf{L}_{\text{given}}[i] / k$\;
    $\bar{f}_{\text{alt}} \gets \mathbf{L}_{\text{alt}}[i] / k$\;
    
    $\Lambda_i \gets \bar{f}_{\text{alt}} / \bar{f}_{\text{given}}$ \tcp*{Likelihood Ratio}
    
    \If{$\Lambda_i > \tau$}{
        Switch binary label of $y_\text{tr}[i]$\;
    }
}
Fine-tune $\mathcal{M}$ with updated labels $y_\text{tr}$ to obtain $\mathcal{M}_\text{adj}$\;
\end{algorithm}

Here we provide a final summary of the ALED algorithm, shown schematically in Algorithm 1. We take the neural network, $\mathbf{f}(\mathbf{x})=\sigma(\mathbf{W}^L\mathbf{x}^L+\mathbf{b}^L)$, trained on a potentially noisy binary classification dataset, $\mathcal{X}$, and remove the final classifier and penultimate activation function, leaving $\mathbf{g}(\mathbf{x})=\mathbf{W}^{L-1}\mathbf{x}^{L-1}+\mathbf{b}^{L-1}$. We use this abbreviated network to compute the set of penultimate feature maps $\mathcal{Z} = \{\mathbf{Z}_1...\}$, each of which we average-pool into vectors, $\mathcal{Z}_{\text{pool}} = \{\mathbf{z}_1...\}$. We partition the data by given class label and calculate means: $\boldsymbol{\mu}_0$ and $\boldsymbol{\mu}_1$ and compute some $r$-dimensional set of approximately orthogonal random projections, $\mathbf{R}=[\mathbf{r}_1 | \mathbf{r}_2 | ... |\mathbf{r}_r]$. We then project the data into a lower-dimensional space specified by $\boldsymbol{\mu}_1 - \boldsymbol{\mu}_0$ and $\mathbf{R}$:
$$
\mathbf{Z}_{\text{reduced}} =  \begin{bmatrix} (\boldsymbol{\mu}_1 - \boldsymbol{\mu}_0)^\text{T } \\ \mathbf{R}^{\text{T }} \end{bmatrix} [\mathbf{z}_1 | \mathbf{z}_2 | ...]
$$
In this lower-dimensional space, we fit Gaussians using the MCD method, estimating $\hat{\boldsymbol{\mu}}_{0}$, $\hat{\boldsymbol{\mu}}_{1}$, $\hat{\boldsymbol{\Sigma}}_{0}$, and $\hat{\boldsymbol{\Sigma}}_{1}$. Given these distributions, we can calculate the likelihoods of belonging to either class for each sample from the multivariate Gaussian probability density function:
$$
f(\mathbf{x}) = \frac{1}{\sqrt{(2\pi)^k |\boldsymbol{\Sigma}|}} \exp\left( -\frac{1}{2} (\mathbf{x} - \boldsymbol{\mu})^\top \boldsymbol{\Sigma}^{-1} (\mathbf{x} - \boldsymbol{\mu}) \right)
$$
We repeat the random projection and likelihood calculation steps for some ensemble size, $k$, and compute average likelihoods, $\bar{f}(\mathbf{z}_{\text{reduced}})$. For each sample, $i$, we have the average computed likelihood of the given label $\bar{f}_{\text{given}}(\mathbf{z}_{\text{reduced}, i})$ and the alternative label $\bar{f}_{\text{alt}}(\mathbf{z}_{\text{reduced}, i})$. We set some likelihood ratio threshold, $\tau$, and we compute the new label $F$ as in Eq. 6. Given $F$, we can identify all of the samples that were putatively mislabeled in the original data. We can also use Bayes' rule (if given a prior distribution) to calculate probabilities of mislabeling.

\section{Experimental Validation}
\subsection{Methods}
We implement ALED in Python, taking advantage of classes and methods from the \texttt{NumPy} \cite{harris2020array}, \texttt{SciPy} \cite{2020SciPy-NMeth}, \texttt{scikit-learn} \cite{scikit-learn}, \texttt{pandas} \cite{reback2020pandas, mckinney-proc-scipy-2010}, and $\texttt{PyTorch}$ \cite{paszke2019pytorchimperativestylehighperformance} libraries, and we compare to CL and CL w/ features in their default implementations \cite{northcutt2021confident}. The core label error detection portion of the ALED algorithm is implemented in the package \texttt{statlab} (retraining is performed separately). ALED takes as input a dataset with potential mislabeling and a model trained on that dataset. For evaluation, we calculate two different kinds of statistics: classification statistics at a given threshold (F1 score, sensitivity, specificity, positive predictive value (PPV), negative predictive value (NPV)) and overall statistics (AUPRC). Classification statistics are produced by thresholding with the likelihood ratio test while the overall statistics are produced by applying Bayes' theorem to the estimated Gaussian distributions, using the proportions of given class labels as priors. ALED requires three hyperparameters: the number of components to retain after dimensionality reduction ($d$), the number of ensembles of Gaussian estimates ($k$), and the likelihood ratio threshold ($\tau$).

We test ALED, CL, and CL with features on two different model architectures (DenseNet121 and ResNet50) and four different datasets (PneumoniaMNIST, BreastMNIST, RetinaMNIST, and BloodMNIST) \cite{huang2017densely, he2016deep, medmnistv2}. We convert RetinaMNIST (an ordinal regression dataset) into a binary classification set by setting all non-zero labels (which correspond to different diabetic retinopathy severities) to 1. For BloodMNIST (a multi-class dataset), we only consider classes 0 (basophil) and 3 (immature granulocytes), to make a binary classification set. 

For each dataset, we use the large image size version ($224 \times 224$), randomly mislabel 5$\%$ of samples in both the training and validation sets, and train the model on the partially-mislabeled training set for a fixed epoch limit, retaining the model with the best validation set performance. We fine-tune both pretrained and randomly initialized versions of each of the aforementioned models. We use the Adam optimizer with a learning rate of $10^{-4}$, a batch size of 8, and an epoch limit of 16 for pretrained models and 32 for untrained models. We then evaluate mislabeled data detection for all three methods. We implement ALED with the following default hyperparameters: $d=2$, $k=10$, and $\tau=2$. These values were tuned using a separate dataset from those presented (the FracAtlas dataset \cite{abedeen2023fracatlas}) to prevent overfitting of the ALED method and ensure a fair comparison with CL. Also, ResNet presented better performance on that dataset, so single-dataset experiments are performed with ResNet in this paper; however, DenseNet provided slightly better performance in general on the MedMNIST datasets presented here.

We further evaluate the three algorithms over various mislabeling rates on a single dataset (1$\%$, 2$\%$, 5$\%$, 10$\%$, and 20$\%$). We also characterize ALED's behavior over a range of each hyperparameter on a single dataset (dimensionality before MCD: 1, 2, 5, 10, 20; ensembles: 1, 2, 5, 10, 20, 50; likelihood ratio thresholds: 1, 2, 5, 10) in Appendix III.

Finally, we demonstrate the performance gains from using each of the three methods to clean the dataset. We fine-tune the pretrained ResNet model on each dataset as before, with a 10\% mislabeling rate, and then we use each method to remove suspected mislabeled samples from the training and validation sets. Then we reinitialize the model and fine-tune on the cleaned dataset from each method. We consider the performance gains of the model on a clean test set before and after cleaning the training/validation datasets. All experiments except for the dataset cleaning experiment are performed for three trials. The dataset cleaning experiment is performed for fifty trials.

\subsection{Results}
The performance of ALED, CL, and CL with features in correctly identifying mislabeled data under the various experimental conditions are reported in Figs. 1–5. For all figures, the mean is reported with error bars corresponding to the standard error of the mean. We report F1 score and AUPRC in the figures, with detailed metrics for all experiments reported in Appendix II and hyperparameter perturbation experiments reported in Appendix III.

A critical finding in our experiments is the distinct trade-off between sensitivity and precision employed by the different methods. As shown in Figs. 1 and 2, CL often has similar AUPRC to ALED (strong theoretical performance with proper thresholding), but its thresholding methods result in reduced real-world performance. Investigating Tables I and II, we see that this is a result of heavily prioritizing PPV over sensitivity. For instance, on PneumoniaMNIST, CL detects only $42.3\%\pm14.2\%$ of mislabeled samples. In contrast, ALED demonstrates a two-to-three-fold increase in sensitivity (achieving $84.7\%\pm6.8\%$ on the same dataset) while maintaining a comparable PPV, resulting in the considerable F1 score improvements in Figs. 1 and 2.

We also observe a divergence in performance based on the model's initialization state. ALED consistently outperforms CL methods when applied to pretrained networks, but this advantage diminishes when networks are trained from scratch (Figs. 3 and 4).

At a dataset level, all methods/models perform quite poorly on BreastMNIST, though ALED consistently provides higher F1 and AUPRC than both CL methods. BreastMNIST has a comparatively small dataset size ($n=546$ training examples), which likely contributes to this behavior.

Continuing, in Fig. 5, we see ALED maintains its performance advantages over a wide range of realistic mislabeling ranges from $1\%$, indicating rare errors, up to $20\%$, which aligns with higher ranges of inter-observer variability in radiology.

Ultimately, the goal of mislabeling detection is not merely to flag errors, but to improve downstream model performance. Fig. 6 illustrates this practical utility: fine-tuning a model on data cleaned by ALED resulted in a $33.8\%$ reduction in test error. This is nearly triple the improvement provided by CL with features ($13.5\%$) and over five times that of standard CL ($6.2\%$).

\subsection{Discussion}
The results highlight a clear trade-off between sensitivity and precision that has important practical implications for dataset cleaning. In the context of dataset cleaning for deep learning, we argue that in many cases sensitivity is paramount; missing a mislabeled sample (a false negative) allows the model to memorize incorrect data, which is often more detrimental than inadvertently removing a correct sample (a false positive), especially in large datasets where data redundancy exists. The additional samples flagged by ALED, which CL often misses due to its conservative thresholding, appear to be genuinely harmful noisy samples that degrade model generalization.

The observed dependence on model initialization empirically supports our theoretical motivation regarding the feature space geometry. Pretrained networks, which have been exposed to massive datasets like ImageNet, likely possess a more structured, Gaussian-like feature space that ALED's covariance estimation relies upon (we discuss this point in more detail below) \cite{papyan2020prevalence}. Conversely, networks trained from scratch on small, noisy datasets may not develop these stable manifold properties, rendering the Gaussian assumption less effective.

The poor performance across all methods on BreastMNIST further emphasizes the role of data availability. The behavior of ALED on BreastMNIST may suggest that ALED is more robust to low data availability than CL. Likely, there is insufficient data to learn detailed trends, rendering the majority of the feature space useless and increasing the likelihood of classifier block overfitting.

Despite this variability, ALED provides the strongest overall performance and in more meaningful cases. Large-scale pretrained models are a common starting point for new ML tasks, and any multi-class task can be decomposed into several binary tasks, as demonstrated for RetinaMNIST and BloodMNIST. Even in the case of training a model from scratch, ALED's method using optimal model settings (either a pretrained DenseNet or ResNet) can be used as a first step for dataset cleaning, prior to training the model of interest.

The role of label noise in deep learning remains a subject of debate. While some studies suggest that small amounts of structured noise, such as label smoothing, can act as a regularizer to prevent overconfidence \cite{szegedy2016rethinking}, unstructured or ``hard'' label noise presents a significant challenge. As demonstrated by Zhang et al. \cite{zhang2017understanding}, deep networks possess the capacity to brute-force memorize corrupt labels, which distorts the decision boundary and degrades generalization on clean data \cite{arpit2017closer}. Our findings strongly support the latter perspective in the context of practical model development.

There are a number of limitations to ALED in its current implementation, and many of them motivate future work. The major assumption of ALED, near-Gaussian pre-activations, is a useful approximation, but often not satisfied in practice \cite{wolinski2025gaussian}. This appears to be most prevalent as an issue for ALED in the untrained model setting. There is evidence for this from the neural collapse literature, which suggests that large-scale datasets such as ImageNet (used for pre-training of ResNet and DenseNet) encourage the formation of dense, well-separated feature manifolds where intra-class feature distributions are highly concentrated and approximately isotropic, making Gaussian approximations more reasonable \cite{papyan2020prevalence}. Future work could involve using particular initializations that encourage Gaussianity (e.g. \cite{wolinski2025gaussian}) in the untrained case to improve ALED's performance.

Additionally, the current form of ALED requires a binary problem, as the initial projection is between given-class means. This requires decomposing multiclass problems into many one-against-all problems, increasing computational costs. Future work exploring more sophisticated projection methods, such as those in Local Fisher Discriminant Analysis \cite{JMLR:v8:sugiyama07b}, could extend ALED to these cases.

Another area of exploration is how the hyperparameters of model training impact ALED and CL performance. In particular, the choice of model training stage (epoch) at which to apply ALED/CL could be explored, as the feature space evolves throughout training. Finally, with ALED we simply chose the final feature outputs to constitute the feature space, but an exploration of features at various depths or combinations of different depth features could provide potential improvements.

\subsection{Conclusion}
We introduce Adaptive Label Error Detection (ALED), a novel framework for mislabeled data detection with DCNNs that shows strong performance in simulated mislabeling scenarios, and which is available as a Python package for easy adoption. ALED demonstrates that exploiting deep feature geometry enables reliable identification of harmful label noise and can translate directly into meaningful improvements in model generalization, highlighting the central role of data quality in deep learning systems.

\begin{figure}[htbp]
    \centering
    \subfloat[F1 Score\label{fig1:sub1}]{%
        \includegraphics[width=0.9\linewidth]{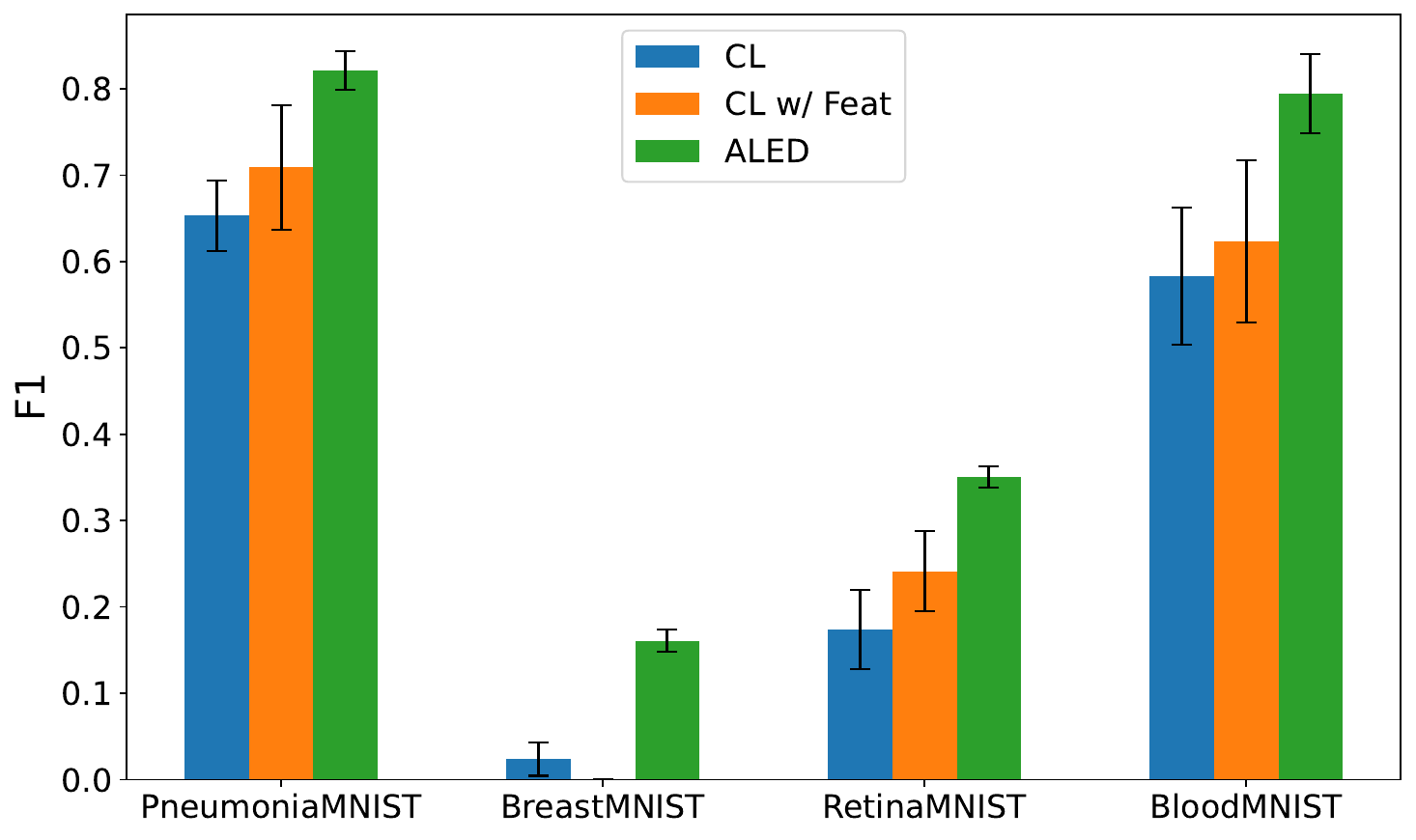}
    }\par\medskip
    \subfloat[AUPRC\label{fig1:sub2}]{%
        \includegraphics[width=0.9\linewidth]{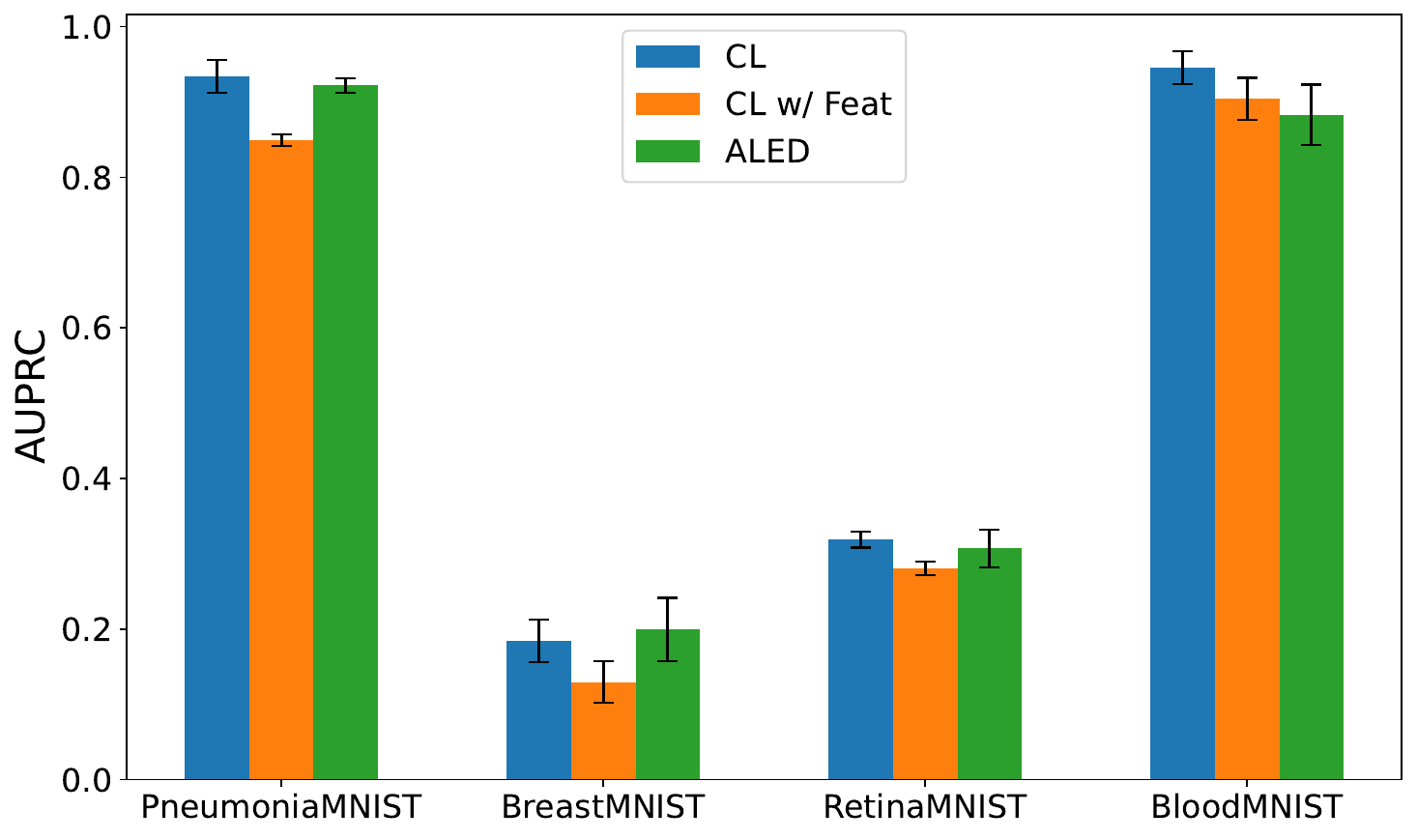}
    }
    \caption{Mislabeled data detection using pretrained DenseNet121 trained on 5$\%$ mislabeled data over all four datasets.}
    \label{fig1:main}
\end{figure}

\begin{figure}[htbp]
    \centering
    \subfloat[F1 Score\label{fig2:sub1}]{%
        \includegraphics[width=0.9\linewidth]{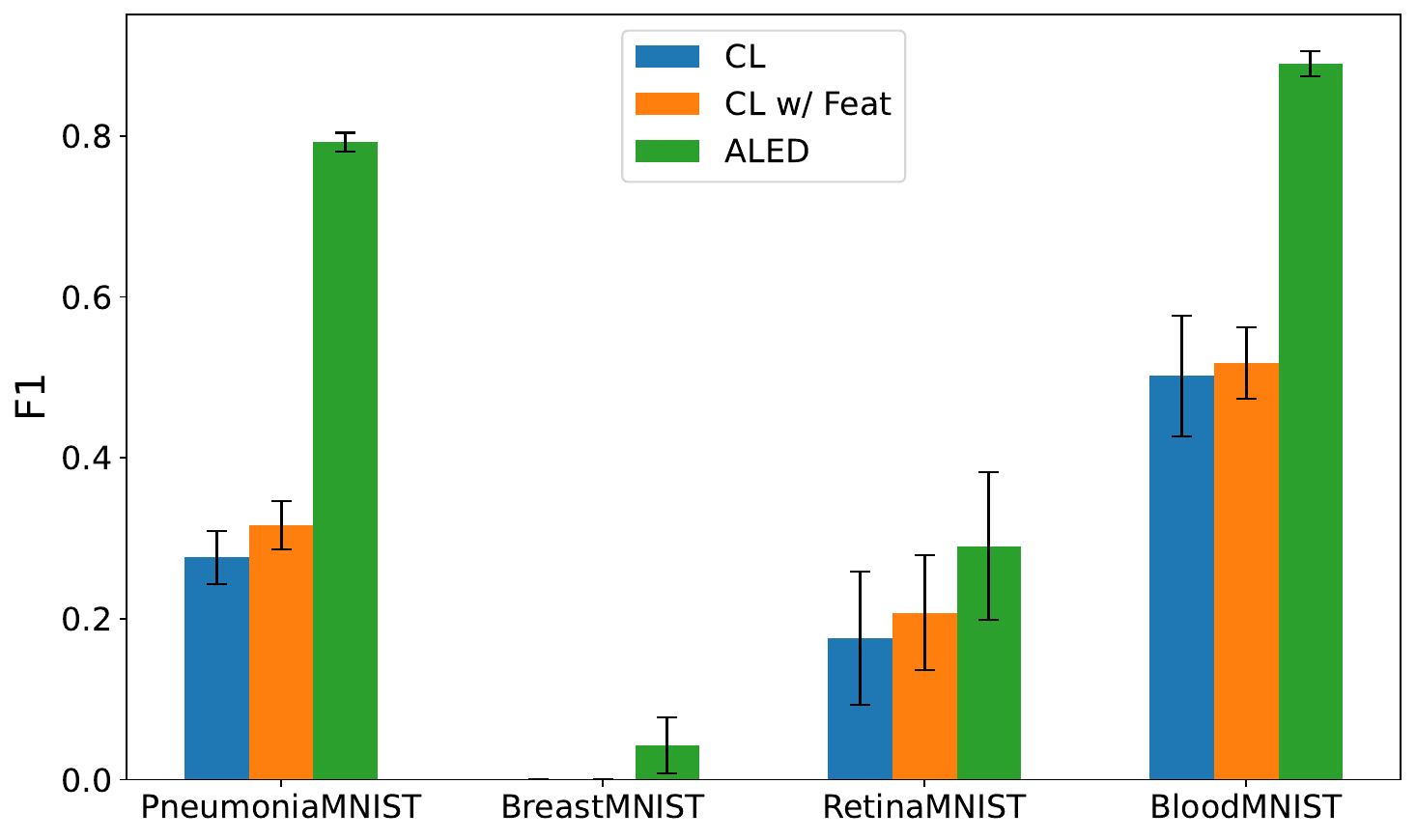}
    }\par\medskip
    \subfloat[AUPRC\label{fig2:sub2}]{%
        \includegraphics[width=0.9\linewidth]{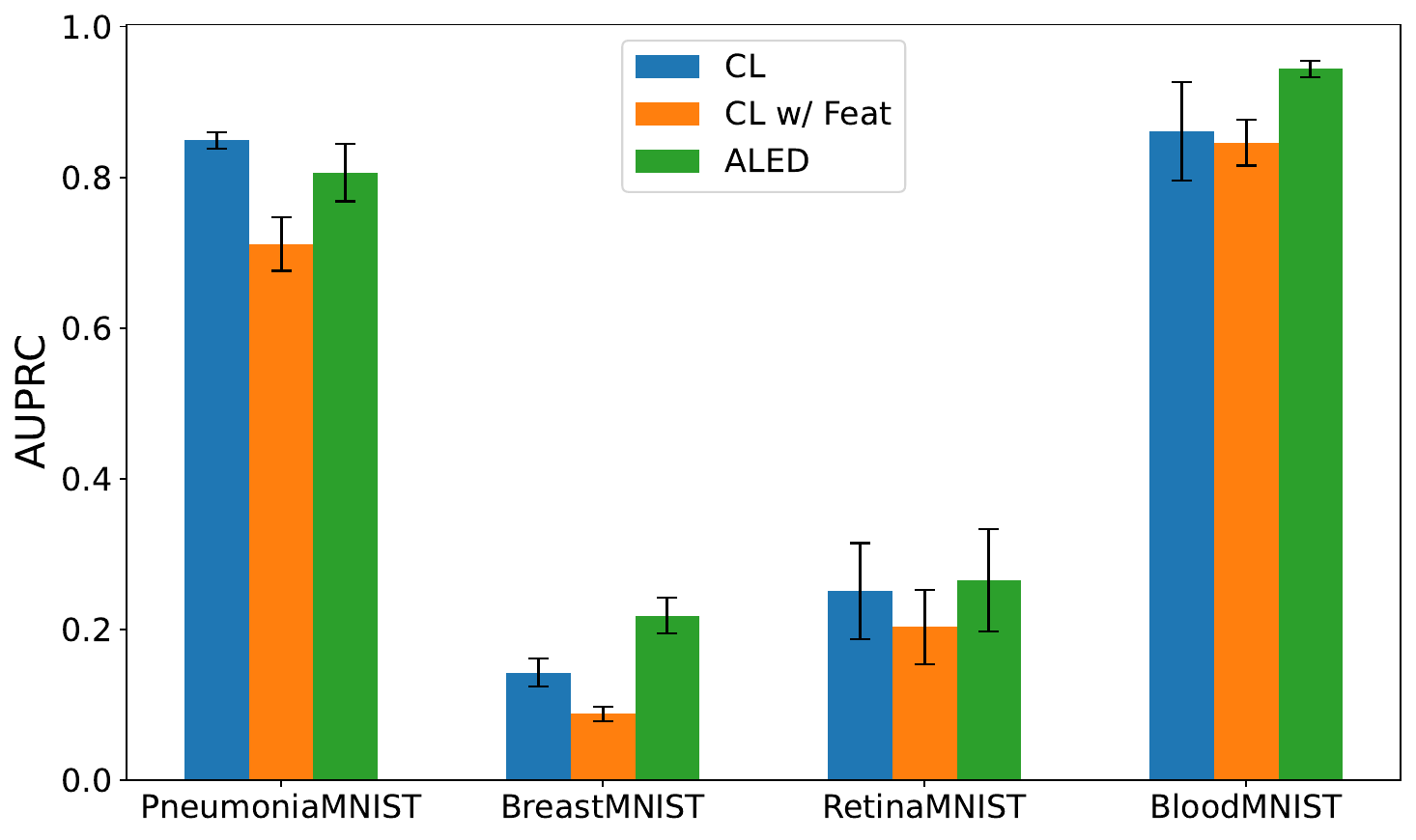}
    }
    \caption{Mislabeled data detection using pretrained ResNet50 trained on 5$\%$ mislabeled data over all four datasets.}
    \label{fig2:main}
\end{figure}

\begin{figure}[htbp]
    \centering
    \subfloat[F1 Score\label{fig3:sub1}]{%
        \includegraphics[width=0.9\linewidth]{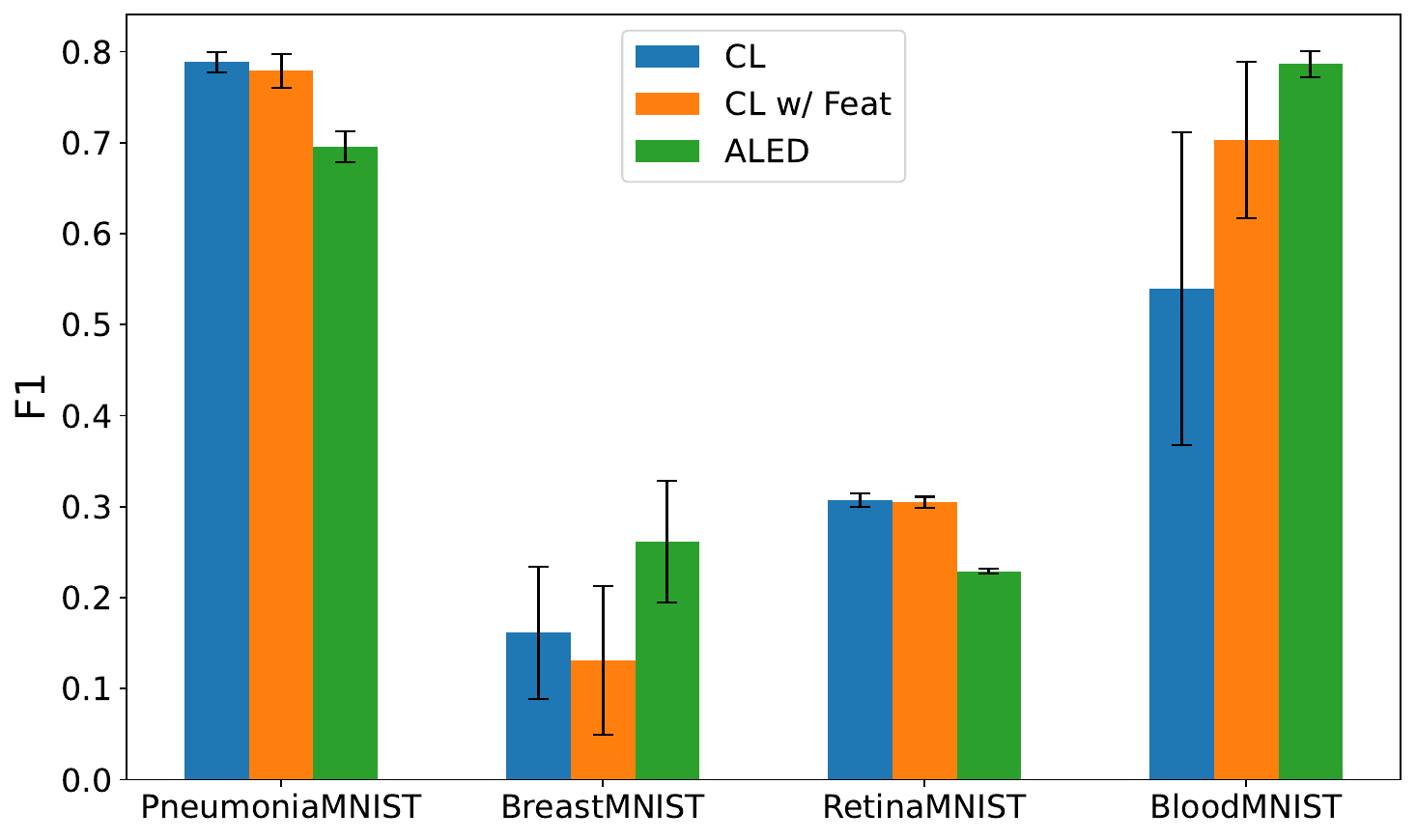}
    }\par\medskip
    \subfloat[AUPRC\label{fig3:sub2}]{%
        \includegraphics[width=0.9\linewidth]{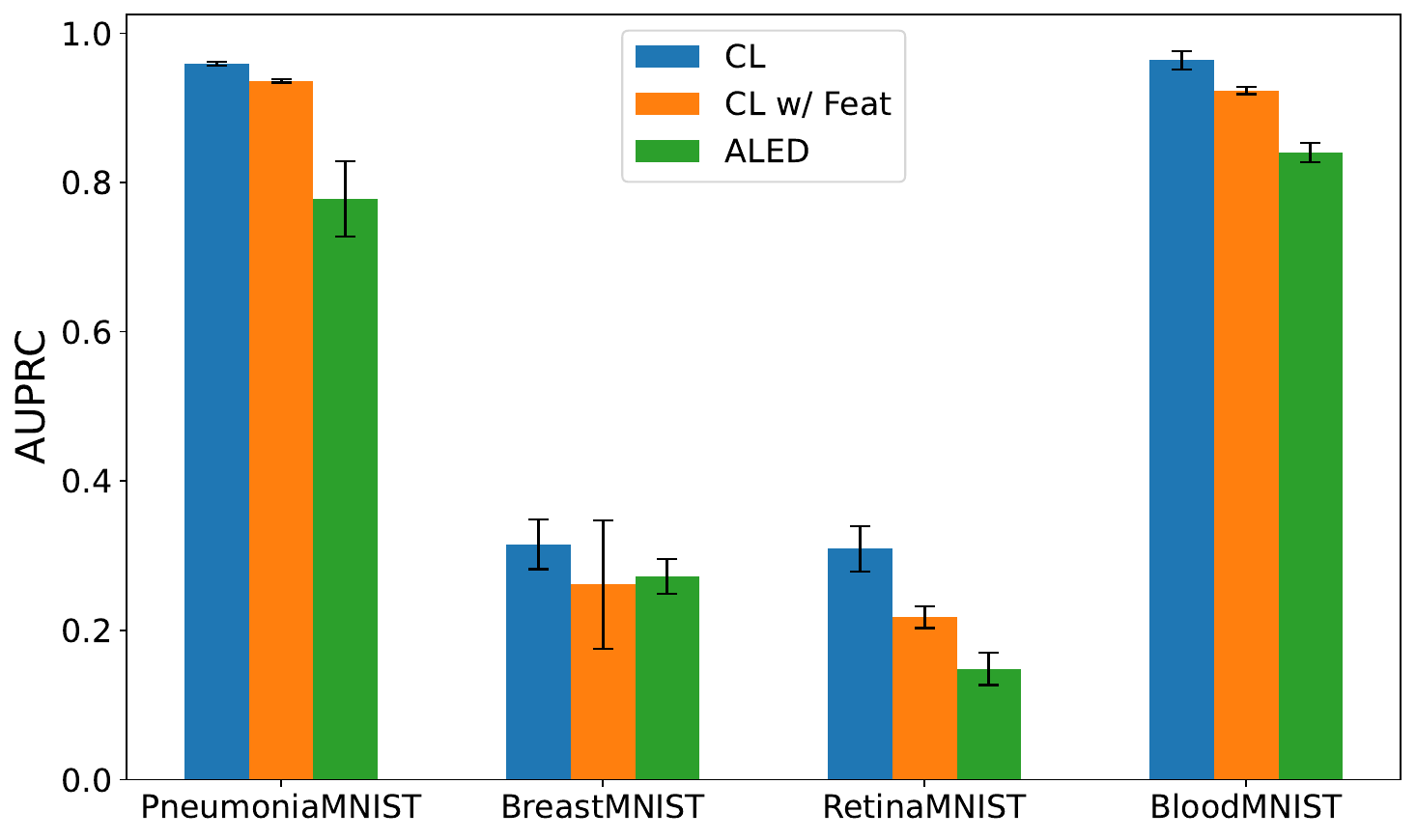}
    }
    \caption{Mislabeled data detection using randomly initialized DenseNet121 trained on 5$\%$ mislabeled data over all four datasets.}
    \label{fig3:main}
\end{figure}

\begin{figure}[htbp]
    \centering
    \subfloat[F1 Score\label{fig3:sub1}]{%
        \includegraphics[width=0.9\linewidth]{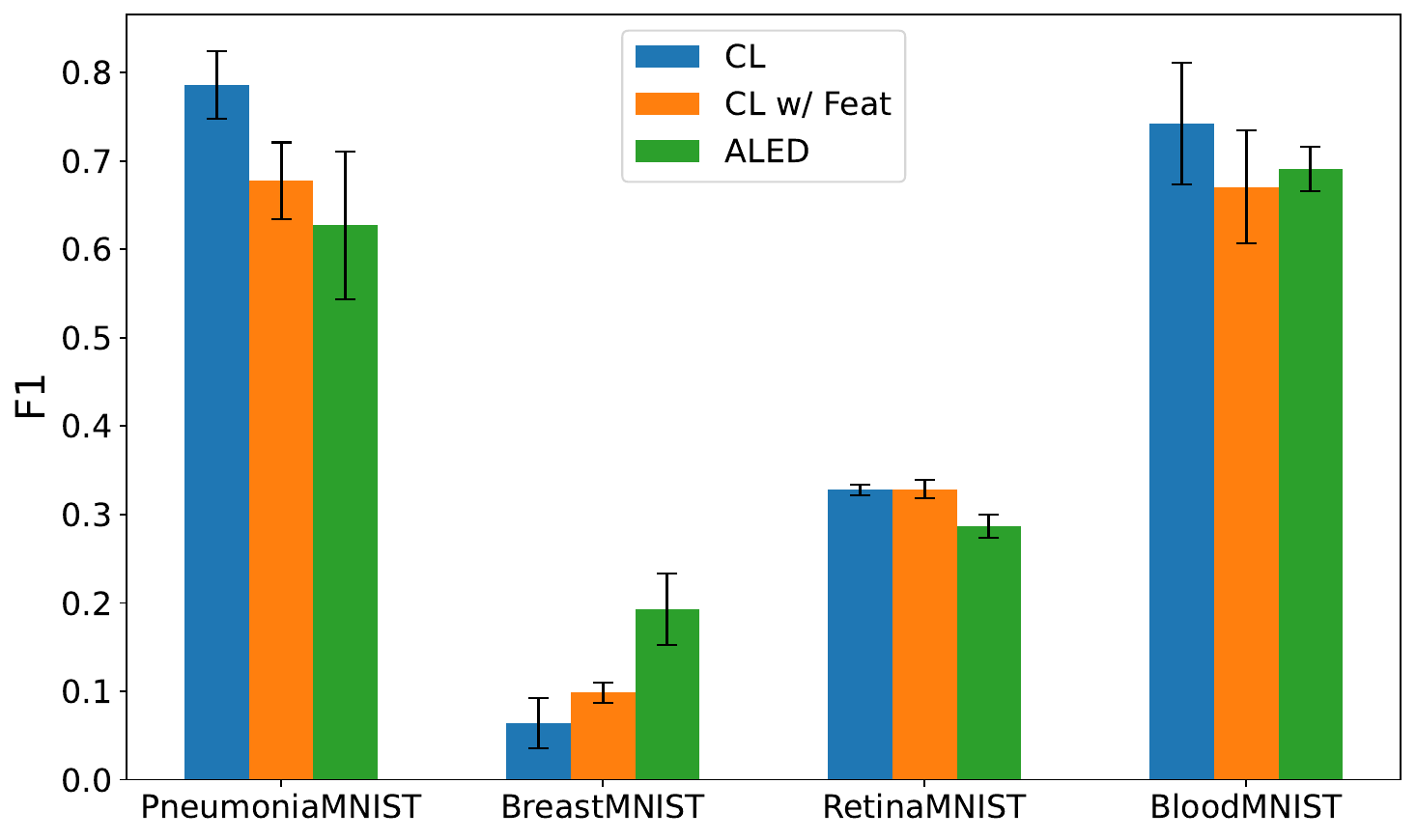}
    }\par\medskip
    \subfloat[AUPRC\label{fig3:sub2}]{%
        \includegraphics[width=0.9\linewidth]{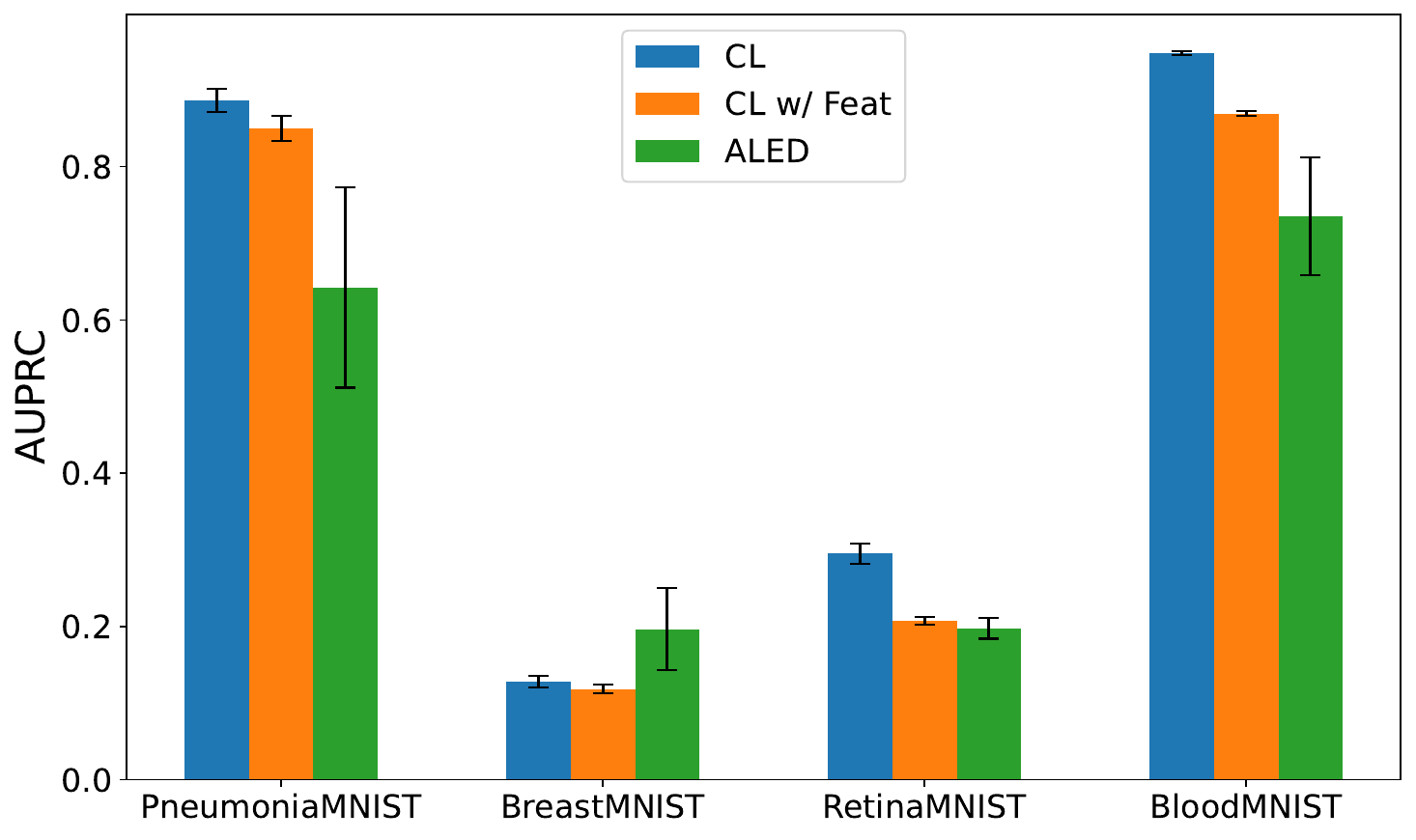}
    }
    \caption{Mislabeled data detection using randomly initialized ResNet50 trained on 5$\%$ mislabeled data over all four datasets.}
    \label{fig3:main}
\end{figure}

\begin{figure}[htbp]
    \centering
    \subfloat[F1 Score over Mislabeling Fraction\label{fig5:sub1}]{%
        \includegraphics[width=0.9\linewidth]{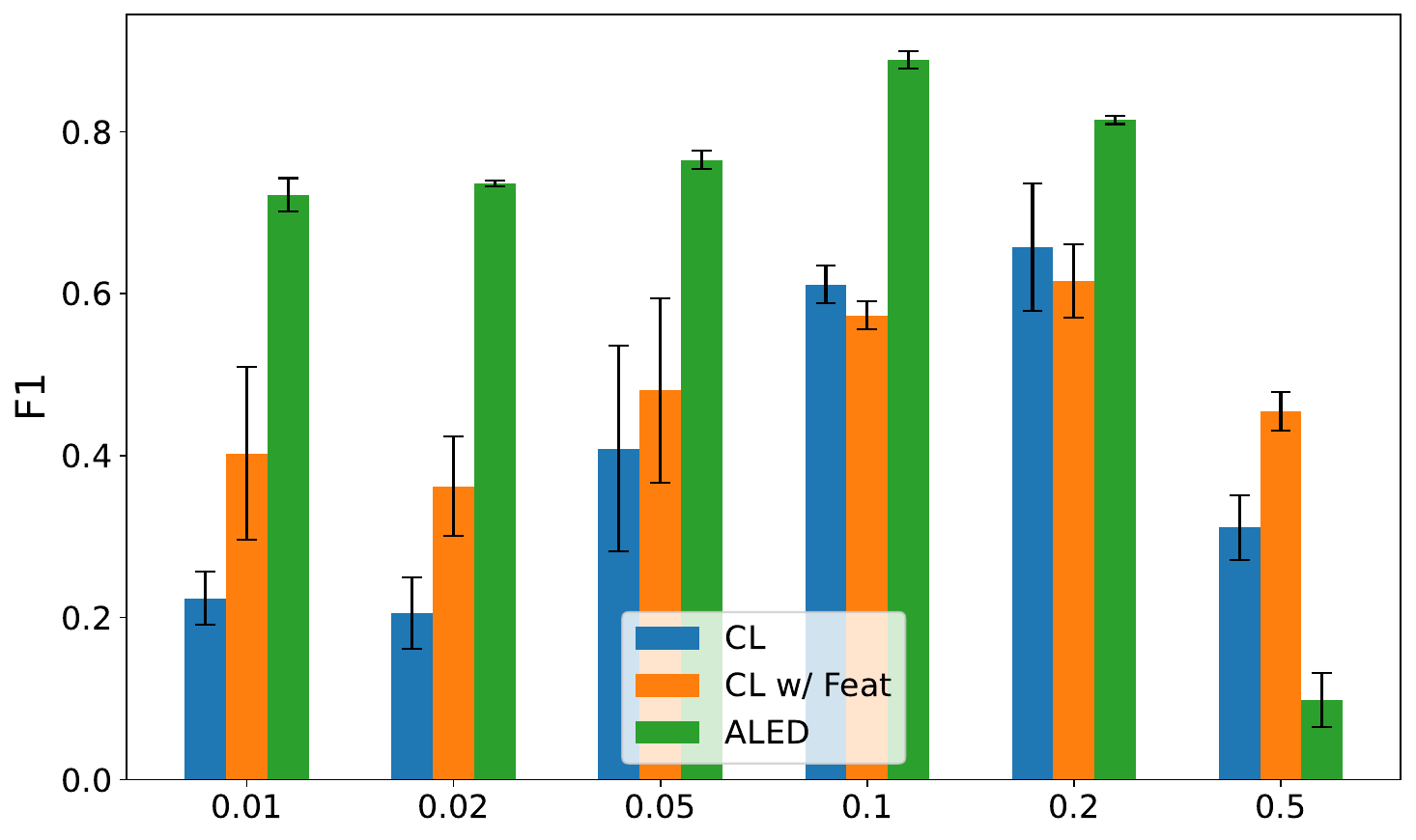}
    }\par\medskip
    \subfloat[AUPRC over Mislabeling Fraction\label{fig5:sub2}]{%
        \includegraphics[width=0.9\linewidth]{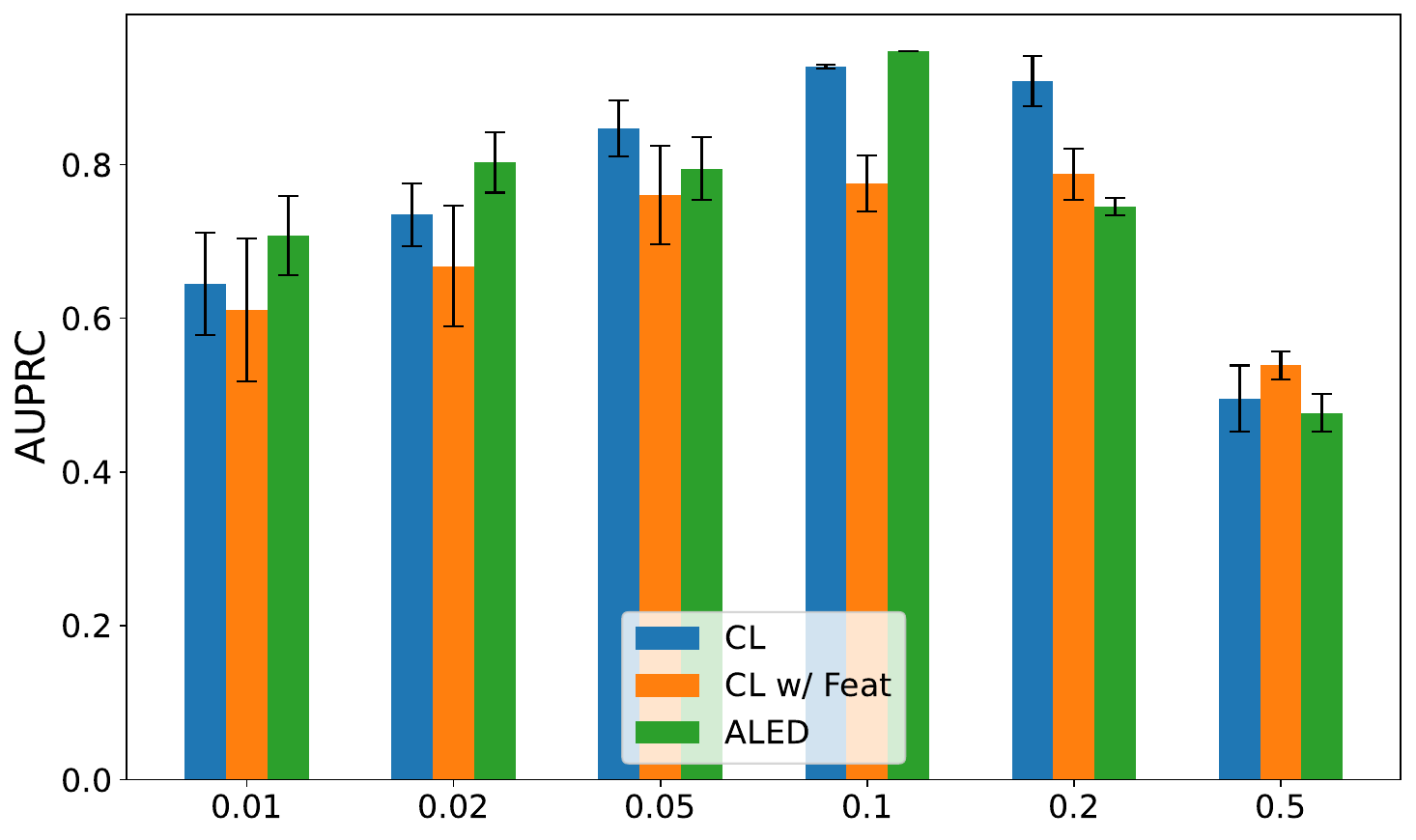}
    }
    \caption{Mislabeled data detection on PneumoniaMNIST using pretrained ResNet50 trained over a range of mislabeling rates.}
    \label{fig5:main}
\end{figure}

\begin{figure}[htbp]
    \centering
    \includegraphics[width=0.9\linewidth]{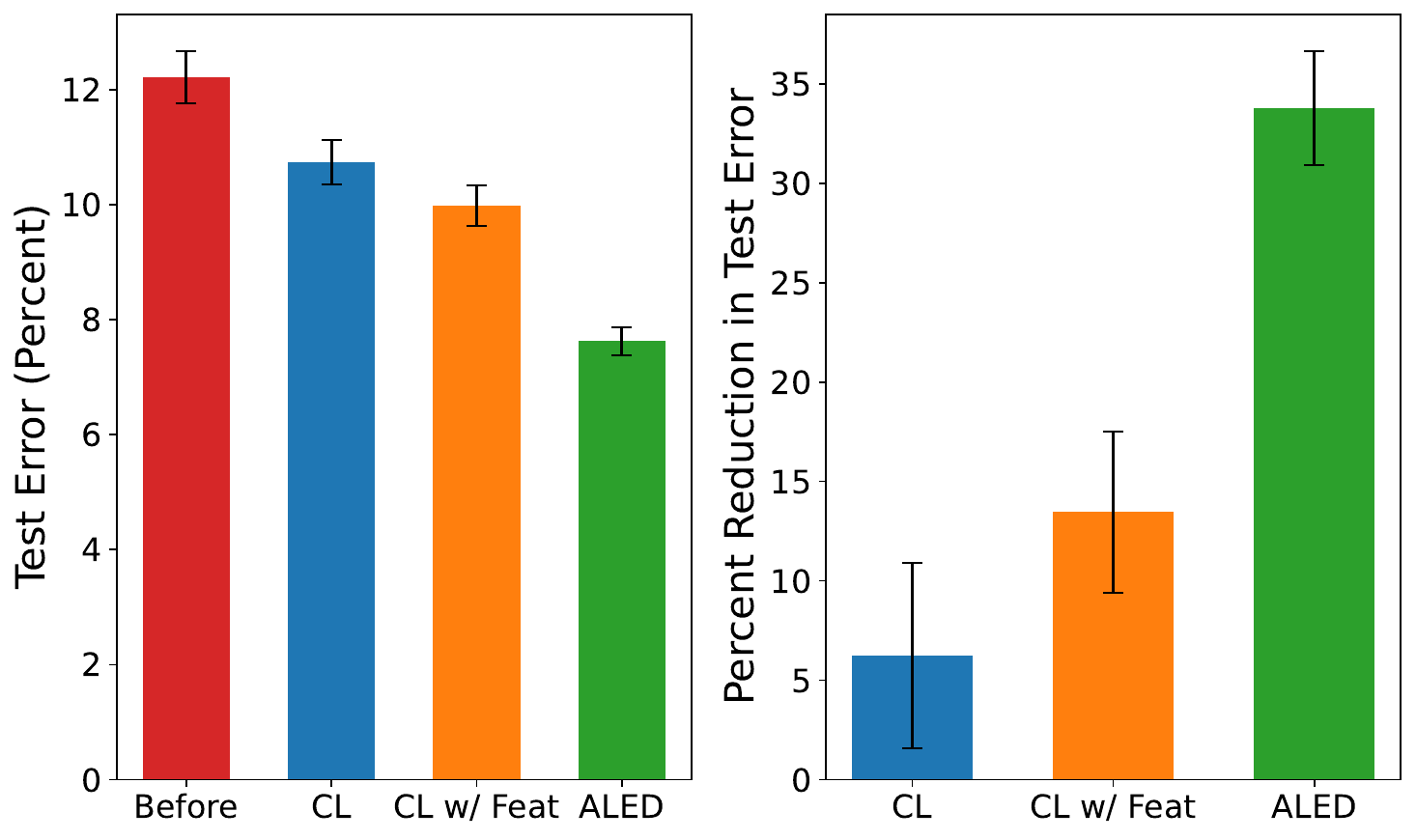}
    \caption{Test performance of a pretrained ResNet on PneumoniaMNIST with 10$\%$ mislabeling, using each method to clean the dataset. Before refers to the baseline model, without any data cleaning. Test error is given by $100-$accuracy.}
    \label{fig4:main}
\end{figure}

\section*{Acknowledgment}
The authors thank Tej Mehta for helping identify the CL sensitivity issue in biomedical image analysis.

\section*{Code and Data Availability}
The code and code outputs used to generate the results for this study are deposited in Zenodo \cite{chaudhry_2026_18210899}. All other data used in the course of this study is freely available online, and can be downloaded (or accessed) as shown in the code. The core error-detection component of ALED is implemented in the Python package \texttt{statlab}.

\appendices
\section{Projection Along Difference Between Means}
In the feature space for a binary classification task, we consider two multivariate Gaussian distributions
$\Vec{X} \sim \mathcal{N}(\Vec{\mu}_1, \mathbf{\Sigma}_1)$ and
$\Vec{Y} \sim \mathcal{N}(\Vec{\mu}_2, \mathbf{\Sigma}_2)$.
We seek a one-dimensional projection that maximizes class separability, measured via the Hellinger distance.

For two univariate Gaussians $X \sim \mathcal{N}(\mu_1,\sigma_1)$ and
$Y \sim \mathcal{N}(\mu_2,\sigma_2)$, the squared Hellinger distance is
\[
H^2(X,Y)
= 1 - \sqrt{\frac{2\sigma_1\sigma_2}{\sigma_1^2 + \sigma_2^2}}
\exp\!\Bigg(
-\frac{(\mu_1-\mu_2)^2}{4(\sigma_1^2+\sigma_2^2)}
\Bigg).
\]

Let $X = \Vec{x}^{\text T}\Vec{X}$ and $Y = \Vec{x}^{\text T}\Vec{Y}$ for a nonzero projection vector $\Vec{x}$.
Assuming a scaled covariance structure $\mathbf{\Sigma}_2 = \beta \mathbf{\Sigma}_1$
(a relaxed LDA assumption justified for DCNN feature spaces \cite{lee2019robustinferencegenerativeclassifiers}),
the projected variances are
$\sigma_1^2 = \Vec{x}^{\text T}\mathbf{\Sigma}\Vec{x}$ and
$\sigma_2^2 = \beta \Vec{x}^{\text T}\mathbf{\Sigma}\Vec{x}$,
where $\mathbf{\Sigma}=\mathbf{\Sigma}_1$.
Defining $\Vec{v}=\Vec{\mu}_1-\Vec{\mu}_2$, the Hellinger distance reduces to
\[
H^2(X,Y)
= 1 - C(\beta)\exp\!\Bigg(
-\frac{(\Vec{x}^{\text T}\Vec{v})^2}{4(1+\beta)\Vec{x}^{\text T}\mathbf{\Sigma}\Vec{x}}
\Bigg),
\]
where $C(\beta)$ is a positive constant independent of $\Vec{x}$.

Maximizing $H^2$ is therefore equivalent to maximizing the Rayleigh quotient
\begin{equation}
\mathcal{R}(\Vec{x})
= \frac{(\Vec{x}^{\text T}\Vec{v})^2}{\Vec{x}^{\text T}\mathbf{\Sigma}\Vec{x}}.
\end{equation}
The stationary condition yields
\begin{equation}
\Vec{v}
= \Bigg(\frac{\Vec{x}^{\text T}\Vec{v}}{\Vec{x}^{\text T}\mathbf{\Sigma}\Vec{x}}\Bigg)
\mathbf{\Sigma}\Vec{x},
\end{equation}
implying $\Vec{x} \propto \mathbf{\Sigma}^{-1}\Vec{v}$, which recovers the Fisher linear discriminant.

In practice, $\mathbf{\Sigma}$ is not explicitly accessible. Empirically and theoretically,
deep networks tend to learn approximately decorrelated representations
\cite{bengio2013representation}, and in high-dimensional regimes the spectrum of $\mathbf{\Sigma}$
concentrates by the Marčenko–Pastur law \cite{marcenko1967distribution}.
We therefore approximate $\mathbf{\Sigma} \approx \lambda\mathbf{I}$, yielding
\begin{equation}
\Vec{v}
\approx
\Bigg(\frac{\Vec{x}^{\text T}\Vec{v}}{\Vec{x}^{\text T}\Vec{x}}\Bigg)\Vec{x}.
\end{equation}
Thus, any $\Vec{x}$ proportional to $\Vec{v}=\Vec{\mu}_1-\Vec{\mu}_2$ approximately maximizes
the Hellinger distance, motivating projection along the difference between class means.

\section{Detailed Mislabeled Data Detection Results}
See Tables 1-4 below; values are bolded if they are greater than 5\% above the next-best algorithm. Mean of three trials plus/minus standard error of the mean displayed.

\begin{table*}[t]
\centering
\caption{Pretrained DenseNet121 with 5$\%$ Mislabeling}
\begin{tabular}{llllll}
\toprule
 &  & PneumoniaMNIST & BreastMNIST & RetinaMNIST & BloodMNIST \\
\midrule
\multirow[t]{4}{*}{CL} & Sensitivity & 42.3\% ± 14.2\% & 0.0\% ± 0.0\% & 14.2\% ± 6.2\% & 28.9\% ± 8.6\% \\
 & Specificity & 99.8\% ± 0.1\% & 100.0\% ± 0.0\% & 98.6\% ± 0.8\% & 100.0\% ± 0.0\% \\
 & PPV & \textbf{94.2\% ± 2.8\%} & 0.0\% ± 0.0\% & 25.5\% ± 11.3\% & 100.0\% ± 0.0\% \\
 & NPV & 97.1\% ± 0.7\% & 95.1\% ± 0.0\% & 95.6\% ± 0.3\% & 96.4\% ± 0.4\% \\
\cline{1-6}
\multirow[t]{4}{*}{CL w/ Feat} & Sensitivity & 53.3\% ± 14.2\% & 2.5\% ± 2.0\% & 22.8\% ± 8.6\% & 38.2\% ± 10.7\% \\
 & Specificity & 99.2\% ± 0.3\% & 99.7\% ± 0.1\% & 97.6\% ± 1.0\% & 99.7\% ± 0.2\% \\
 & PPV & 79.8\% ± 4.8\% & 13.3\% ± 10.9\% & 31.8\% ± 4.1\% & 92.2\% ± 4.3\% \\
 & NPV & 97.6\% ± 0.7\% & 95.2\% ± 0.1\% & 96.0\% ± 0.4\% & 96.9\% ± 0.5\% \\
\cline{1-6}
\multirow[t]{4}{*}{ALED} & Sensitivity & \textbf{84.7\% ± 6.8\%} & 7.4\% ± 4.6\% & \textbf{46.9\% ± 12.5\%} & \textbf{90.7\% ± 1.8\%} \\
 & Specificity & 99.1\% ± 0.2\% & 99.1\% ± 0.4\% & 93.0\% ± 2.8\% & 99.8\% ± 0.0\% \\
 & PPV & 84.1\% ± 2.2\% & \textbf{28.6\% ± 12.1\%} & 32.2\% ± 5.7\% & 96.3\% ± 0.9\% \\
 & NPV & 99.2\% ± 0.4\% & 95.4\% ± 0.2\% & 97.1\% ± 0.6\% & 99.5\% ± 0.1\% \\
\cline{1-6}
\bottomrule
\end{tabular}
\end{table*}

\begin{table*}[t]
\centering
\caption{Pretrained ResNet50 with 5$\%$ Mislabeling}
\begin{tabular}{llllll}
\toprule
 &  & PneumoniaMNIST & BreastMNIST & RetinaMNIST & BloodMNIST \\
\midrule
\multirow[t]{4}{*}{CL} & Sensitivity & 16.5\% ± 2.3\% & 0.0\% ± 0.0\% & 11.7\% ± 5.7\% & 35.7\% ± 6.9\% \\
 & Specificity & 99.9\% ± 0.0\% & 99.9\% ± 0.1\% & 99.5\% ± 0.2\% & 99.8\% ± 0.1\% \\
 & PPV & \textbf{90.7\% ± 0.9\%} & 0.0\% ± 0.0\% & 36.0\% ± 15.6\% & 90.8\% ± 5.8\% \\
 & NPV & 95.8\% ± 0.1\% & 95.1\% ± 0.0\% & 95.5\% ± 0.3\% & 96.7\% ± 0.3\% \\
\cline{1-6}
\multirow[t]{4}{*}{CL w/ Feat} & Sensitivity & 20.3\% ± 2.5\% & 0.0\% ± 0.0\% & 15.4\% ± 5.6\% & 35.9\% ± 4.0\% \\
 & Specificity & 99.6\% ± 0.1\% & 99.9\% ± 0.1\% & 98.9\% ± 0.4\% & 99.9\% ± 0.0\% \\
 & PPV & 77.6\% ± 5.2\% & 0.0\% ± 0.0\% & \textbf{45.4\% ± 3.1\%} & 94.6\% ± 1.6\% \\
 & NPV & 96.0\% ± 0.1\% & 95.0\% ± 0.0\% & 95.7\% ± 0.3\% & 96.8\% ± 0.2\% \\
\cline{1-6}
\multirow[t]{4}{*}{ALED} & Sensitivity & \textbf{80.4\% ± 2.6\%} & 3.7\% ± 3.0\% & \textbf{36.4\% ± 13.5\%} & \textbf{87.6\% ± 1.3\%} \\
 & Specificity & 98.8\% ± 0.4\% & 98.8\% ± 0.8\% & 95.9\% ± 1.7\% & 99.5\% ± 0.1\% \\
 & PPV & 79.2\% ± 5.3\% & \textbf{5.0\% ± 4.1\%} & 33.3\% ± 2.4\% & 90.4\% ± 1.8\% \\
 & NPV & 99.0\% ± 0.1\% & 95.2\% ± 0.1\% & 96.7\% ± 0.6\% & 99.4\% ± 0.1\% \\
\cline{1-6}
\bottomrule
\end{tabular}
\end{table*}

\begin{table*}[t]
\centering
\caption{Randomly Initialized DenseNet121 with 5$\%$ Mislabeling}
\begin{tabular}{llllll}
\toprule
 &  & PneumoniaMNIST & BreastMNIST & RetinaMNIST & BloodMNIST \\
\midrule
\multirow[t]{4}{*}{CL} & Sensitivity & 67.1\% ± 2.0\% & 13.6\% ± 7.1\% & 59.3\% ± 6.6\% & 43.8\% ± 18.4\% \\
 & Specificity & 99.8\% ± 0.0\% & 98.7\% ± 0.8\% & 88.1\% ± 1.6\% & 100.0\% ± 0.0\% \\
 & PPV & \textbf{95.8\% ± 0.9\%} & 25.0\% ± 10.5\% & 21.0\% ± 0.5\% & 99.2\% ± 0.6\% \\
 & NPV & 98.3\% ± 0.1\% & 95.7\% ± 0.3\% & 97.6\% ± 0.3\% & 97.2\% ± 0.9\% \\
\cline{1-6}
\multirow[t]{4}{*}{CL w/ Feat} & Sensitivity & 71.6\% ± 3.1\% & 12.3\% ± 8.6\% & 66.7\% ± 4.6\% & 59.0\% ± 11.8\% \\
 & Specificity & 99.4\% ± 0.0\% & 98.6\% ± 0.9\% & 85.8\% ± 0.9\% & 99.8\% ± 0.1\% \\
 & PPV & 85.6\% ± 0.3\% & 19.0\% ± 8.0\% & 19.8\% ± 0.2\% & 94.4\% ± 1.1\% \\
 & NPV & 98.5\% ± 0.2\% & 95.6\% ± 0.4\% & 98.0\% ± 0.3\% & 97.9\% ± 0.6\% \\
\cline{1-6}
\multirow[t]{4}{*}{ALED} & Sensitivity & \textbf{89.6\% ± 1.9\%} & \textbf{29.6\% ± 12.6\%} & 66.0\% ± 1.8\% & \textbf{82.5\% ± 7.3\%} \\
 & Specificity & 96.4\% ± 0.2\% & 96.6\% ± 1.5\% & 78.3\% ± 0.8\% & 98.6\% ± 0.4\% \\
 & PPV & 56.8\% ± 1.6\% & \textbf{36.2\% ± 5.7\%} & 13.8\% ± 0.2\% & 77.5\% ± 4.7\% \\
 & NPV & 99.4\% ± 0.1\% & 96.4\% ± 0.6\% & 97.8\% ± 0.1\% & 99.1\% ± 0.4\% \\
\cline{1-6}
\bottomrule
\end{tabular}
\end{table*}

\begin{table*}[t]
\centering
\caption{Randomly Initialized ResNet50 with 5$\%$ Mislabeling}
\begin{tabular}{llllll}
\toprule
 &  & PneumoniaMNIST & BreastMNIST & RetinaMNIST & BloodMNIST \\
\midrule
\multirow[t]{4}{*}{CL} & Sensitivity & 70.4\% ± 6.3\% & 4.9\% ± 2.7\% & 61.1\% ± 9.7\% & 63.9\% ± 9.9\% \\
 & Specificity & 99.6\% ± 0.1\% & 98.7\% ± 1.0\% & 89.0\% ± 2.4\% & 99.7\% ± 0.1\% \\
 & PPV & \textbf{90.7\% ± 1.1\%} & 37.7\% ± 25.6\% & 23.4\% ± 1.3\% & \textbf{94.2\% ± 2.6\%} \\
 & NPV & 98.5\% ± 0.3\% & 95.2\% ± 0.1\% & 97.8\% ± 0.5\% & 98.2\% ± 0.5\% \\
\cline{1-6}
\multirow[t]{4}{*}{CL w/ Feat} & Sensitivity & 59.6\% ± 6.8\% & 7.4\% ± 1.7\% & 63.6\% ± 4.8\% & 58.5\% ± 9.5\% \\
 & Specificity & 99.2\% ± 0.2\% & 98.0\% ± 1.2\% & 88.3\% ± 1.1\% & 99.4\% ± 0.2\% \\
 & PPV & 81.5\% ± 2.8\% & \textbf{45.2\% ± 22.6\%} & 22.3\% ± 0.8\% & 84.8\% ± 3.3\% \\
 & NPV & 97.9\% ± 0.3\% & 95.3\% ± 0.0\% & 97.9\% ± 0.2\% & 97.9\% ± 0.5\% \\
\cline{1-6}
\multirow[t]{4}{*}{ALED} & Sensitivity & \textbf{86.8\% ± 0.7\%} & \textbf{17.3\% ± 3.6\%} & \textbf{71.6\% ± 5.8\%} & \textbf{87.2\% ± 3.7\%} \\
 & Specificity & 94.4\% ± 2.1\% & 96.1\% ± 1.6\% & 82.4\% ± 2.7\% & 96.6\% ± 0.3\% \\
 & PPV & 51.1\% ± 10.1\% & 36.4\% ± 18.0\% & 18.2\% ± 1.4\% & 57.4\% ± 2.3\% \\
 & NPV & 99.3\% ± 0.1\% & 95.7\% ± 0.2\% & 98.2\% ± 0.3\% & 99.3\% ± 0.2\% \\
\cline{1-6}
\bottomrule
\end{tabular}
\end{table*}

\section{ALED Hyperparameter Grid Search Results}
See Tables 5-7 below. Experiments are performed using pretrained ResNet50 on PneumoniaMNIST with 5\% mislabeling. One model instance was trained and used for all conditions, with ALED repeated for three trials at each combination. Mean plus/minus standard error of the mean displayed.

\begin{table*}[htbp]
\centering
\caption{Varying ALED Projection Number, Fixed Ensemble Number (10) and Likelihood Ratio (2)}
\begin{tabular}{lllllll}
\toprule
 & 1 & 2 & 5 & 10 & 20 & 50 \\
\midrule
Sensitivity & 86.4\% ± 0.0\% & 86.8\% ± 0.0\% & 87.1\% ± 0.1\% & 87.7\% ± 0.0\% & 87.7\% ± 0.0\% & 87.4\% ± 0.4\% \\
Specificity & 99.0\% ± 0.0\% & 99.1\% ± 0.0\% & 99.1\% ± 0.0\% & 99.0\% ± 0.0\% & 99.1\% ± 0.0\% & 99.3\% ± 0.0\% \\
PPV & 82.5\% ± 0.0\% & 83.7\% ± 0.1\% & 82.9\% ± 0.2\% & 82.5\% ± 0.4\% & 83.3\% ± 0.6\% & 87.5\% ± 0.3\% \\
NPV & 99.3\% ± 0.0\% & 99.3\% ± 0.0\% & 99.3\% ± 0.0\% & 99.3\% ± 0.0\% & 99.3\% ± 0.0\% & 99.3\% ± 0.0\% \\
AUPRC & 86.5\% ± 0.0\% & 88.3\% ± 0.2\% & 91.9\% ± 0.1\% & 91.4\% ± 0.1\% & 90.3\% ± 0.5\% & 89.5\% ± 0.3\% \\
\bottomrule
\end{tabular}
\end{table*}

\begin{table*}[htbp]
\centering
\caption{Varying ALED Number of Ensembles, Fixed Projection Number (2) and Likelihood Ratio (2)} 
\begin{tabular}{lllllll}
\toprule
 & 1 & 2 & 5 & 10 & 20 & 50 \\
\midrule
Sensitivity & 87.4\% ± 0.4\% & 87.0\% ± 0.1\% & 87.1\% ± 0.1\% & 86.8\% ± 0.0\% & 86.7\% ± 0.1\% & 86.8\% ± 0.0\% \\
Specificity & 99.1\% ± 0.1\% & 99.1\% ± 0.0\% & 99.1\% ± 0.0\% & 99.1\% ± 0.0\% & 99.1\% ± 0.0\% & 99.1\% ± 0.0\% \\
PPV & 83.4\% ± 1.2\% & 83.2\% ± 0.2\% & 83.7\% ± 0.4\% & 83.7\% ± 0.1\% & 83.2\% ± 0.0\% & 83.4\% ± 0.1\% \\
NPV & 99.3\% ± 0.0\% & 99.3\% ± 0.0\% & 99.3\% ± 0.0\% & 99.3\% ± 0.0\% & 99.3\% ± 0.0\% & 99.3\% ± 0.0\% \\
AUPRC & 89.2\% ± 0.8\% & 88.3\% ± 0.3\% & 88.8\% ± 0.4\% & 88.3\% ± 0.2\% & 88.3\% ± 0.1\% & 88.5\% ± 0.0\% \\
\bottomrule
\end{tabular}
\end{table*}

\begin{table*}
\centering
\caption{Varying ALED Likelihood Ratio, Fixed Projection Number (2) and Ensemble Number (10)}
\begin{tabular}{lllllll}
\toprule
 & 1 & 2 & 5 & 10 & 20 & 50 \\
\midrule
Sensitivity & 87.7\% ± 0.2\% & 86.8\% ± 0.0\% & 86.0\% ± 0.0\% & 85.4\% ± 0.1\% & 84.1\% ± 0.1\% & 83.8\% ± 0.0\% \\
Specificity & 99.1\% ± 0.0\% & 99.1\% ± 0.0\% & 99.2\% ± 0.0\% & 99.2\% ± 0.0\% & 99.3\% ± 0.0\% & 99.3\% ± 0.0\% \\
PPV & 83.1\% ± 0.2\% & 83.7\% ± 0.1\% & 84.4\% ± 0.4\% & 85.3\% ± 0.1\% & 85.8\% ± 0.1\% & 86.7\% ± 0.1\% \\
NPV & 99.3\% ± 0.0\% & 99.3\% ± 0.0\% & 99.3\% ± 0.0\% & 99.2\% ± 0.0\% & 99.2\% ± 0.0\% & 99.2\% ± 0.0\% \\
AUPRC & 88.7\% ± 0.2\% & 88.3\% ± 0.2\% & 88.4\% ± 0.1\% & 88.5\% ± 0.0\% & 88.4\% ± 0.1\% & 88.1\% ± 0.1\% \\
\bottomrule
\end{tabular}
\end{table*}

\bibliographystyle{IEEEtran}
\bibliography{bibtex/bib/Noam_bib}

\end{document}